\documentclass[11pt]{article}
\usepackage{coling2020}
\usepackage{times}
\usepackage{url}
\usepackage{latexsym}

\usepackage{color}

\usepackage{hyperref}

\usepackage[inline]{enumitem}
\usepackage{scrextend}
\usepackage{array}
\usepackage{arydshln}
\usepackage{graphicx}
\usepackage{hhline}

\usepackage{amsmath}

\usepackage{arydshln}
\usepackage{array}
\newcolumntype{L}[1]{>{\raggedright\let\newline\\\arraybackslash\hspace{0pt}}m{#1}}
\newcolumntype{C}[1]{>{\centering\let\newline\\\arraybackslash\hspace{0pt}}m{#1}}
\newcolumntype{R}[1]{>{\raggedleft\let\newline\\\arraybackslash\hspace{0pt}}m{#1}}

\colingfinalcopy 

\title{CxGBERT: BERT meets Construction Grammar}

\author{
Harish Tayyar Madabushi\textsuperscript{1} \and
Laurence Romain\textsuperscript{2} \and
\\
\textbf{Dagmar Divjak\textsuperscript{2,3}} \and
\textbf{Petar Milin\textsuperscript{2}}
\\[0.3cm]
\textsuperscript{1} School of Computer Science \\
\textsuperscript{2} Department of Modern Languages\\
\textsuperscript{3} Department of English Language and Linguistics \\
University of Birmingham, UK\\[3mm]
\texttt{\small Harish@HarishTayyarMadabushi.com} \\[-0.1mm]
\texttt{\small(L.M.Y.Romain, D.Divjak, P.Milin)@bham.ac.uk}\\
\\
}

\date{}

\begin{document}
\maketitle
\begin{abstract}
 While lexico-semantic elements no doubt capture a large amount of linguistic information, it has been argued that they do not capture all information contained in text. This assumption is central to constructionist approaches to language which argue that language consists of \textit{constructions}, learned pairings of a form and a function or meaning that are either frequent or have a meaning that cannot be predicted from its component parts. BERT’s training objectives give it access to a tremendous amount of lexico-semantic information, and while BERTology has shown that BERT captures certain important linguistic dimensions, there have been no studies exploring the extent to which BERT might have access to constructional information. In this work we design several probes and conduct extensive experiments to answer this question. Our results allow us to conclude that BERT does indeed have access to a significant amount of information, much of which linguists typically call constructional information. The impact of this observation is potentially far-reaching as it provides insights into what deep learning methods learn from text, while also showing that information contained in constructions is redundantly encoded in lexico-semantics.
\end{abstract}

\section{Introduction and Motivation}

%
%
\blfootnote{
    %
    %
    %
    %
    \hspace{-0.65cm}  
    Accepted for Publication at the 28\textsuperscript{th} International Conference on Computational Linguistics (COLING 2020) \\
    \hspace{-0.65cm}  
    This work is licensed under a Creative Commons 
    Attribution 4.0 International Licence.
    Licence details:
    \url{http://creativecommons.org/licenses/by/4.0/}.
    %
    %
}

The introduction of pre-trained contextual word embeddings has had a tremendous impact on Natural Language Processing (NLP), resulting in significant improvements on several tasks such as translation and question answering, where the best automated systems are at par with or outperform humans \cite{DBLP:journals/corr/abs-1803-05567}. These models can broadly be classified based on the approach they use: the pre-training approach and the feature-based approach. While methods that rely on features, such as ELMo~\cite{DBLP:journals/corr/abs-1802-05365} require task specific architectures, those that rely on fine-tuning, such as Generative Pre-trained Transformer (OpenAI GPT) \cite{radford2018improving}, update all parameters during fine-tuning thus transferring their learning to new tasks with minimal fine-tuning. While not the first pre-trained model, BERT \cite{DBLP:journals/corr/abs-1810-04805} was the first extremely successful one, utilising a deep bidirectional transformer model to provide significant improvements on the state-of-the-art of several tasks.

The success of BERT and its effectiveness in transfer learning, which is inherent to pre-trained models, could imply that these models have an understanding of the underlying language. This has led several researchers to focus on explaining what it is that these models understand about language. Despite the inherent difficulty in understanding deep learning models, a new subfield of NLP, called BERTology~\cite{DBLP:journals/corr/abs-2002-12327}, has evolved to better understand what BERT captures within its structure and how it is able to so effectively transfer that knowledge to so many different tasks through relatively quick fine-tuning. These efforts have shown that the BERT embeddings capture linguistic information such as tense, parts of speech~\cite{chrupala-alishahi-2019-correlating,tenney2018what}, entire parse trees~\cite{hewitt-manning-2019-structural} in addition to encapsulating the NLP pipeline~\cite{DBLP:journals/corr/abs-1905-05950}. 

BERT is typically trained on two objectives: the Masked Language Model (MLM) pre-training objective (which involves randomly masking some tokens and setting the objective to predict the original tokens), and Next Sentence Prediction (NSP) objective (which consists of predicting if, given two sentences, one follows the other or not). These training objectives provide BERT with access to a tremendous amount of lexico-semantic information.

While lexico-semantic elements no doubt capture a large amount of linguistic information, it has been argued that they do not capture all information contained in text~\cite{goldberg1995constructions}. This assumption is central to the constructionist approach which does not adhere to a strict division between lexical and grammatical elements, but rather argues that they form a continuum made up of  \textit{constructions}, which are learned pairings of form and function or meaning. Constructions include partially or fully filled words, idioms or general linguistic patterns (e.g. idiom, partially filled:  \texttt{believe <one’s> ears/eyes}). In order for a construction to be considered as such, its meaning must not be predictable from the sum of its parts (for example, in the case of idioms), or alternatively, if this meaning is predictable then a pattern is considered a construction if it occurs frequently~\cite{goldberg2006constructions}. Constructions occur at multiple levels of linguistic organisation (e.g., morphology, syntax) and at different levels of generalisation, starting with low level constructions, such as \texttt{cat + -s} (i.e., cats), progressing to more complex constructions such as \texttt{Noun + -s} (the plural construction) or \texttt{The Xer the Yer} (e.g., The more I think about it, the less I like it), and finally the most abstract level which no longer contains any lexical elements such as the \texttt{ditransitive construction} (e.g. She gave me a book; I emailed her the details).

Speakers of a language are capable of making generalisations based on their encounter with only a few instances of a construction~\cite{hilpert2014construction}. The constructionist approach is based on theoretical principles developed through the study of language learners: it encompasses both the study of constructions (Construction Grammar: CxG, see Section \ref{section:cxg}) and the idea that language learners construct their knowledge of language based on input (what people hear and read), subject to cognitive and practical constraints~\cite{goldberg2006constructions}.

While BERTology has shown that BERT captures certain important linguistic information, there have been no studies exploring the extent to which BERT has access to constructional information. We hypothesise that should BERT have access to constructional information, it should perform well on distinguishing constructions from each other. However, the ability to do so is not sufficient to conclude that BERT has constructional information as we must fine-tune BERT for it to perform this operation. Additionally, BERT could be using other information to distinguish constructions. If BERT lacks constructional information, explicitly adding this information will significantly alter BERT’s ability to perform on downstream tasks or its ability to represent elements of language such as PoS or parse trees, especially given the results of linguistic and cognitive linguistic experiments (Sections \ref{section:cxg} and \ref{section:cxg-semantics}). On the other hand, should BERT already contain constructional information, the addition of this information would have no significant effect thus showing that BERT  does contain constructional information, or information that is functionally equivalent. Thus, this work focuses on answering the following questions:
\begin{enumerate}[label=(\alph*),leftmargin=4em]
    \itemsep 0em 
    \item How does the addition of constructional information affect BERT? 
    \item How effective is BERT in identifying constructions?
\end{enumerate}
    
\textbf{ Answers to these questions will enable us to determine the extent to which BERT has access to constructional information.} It is crucial that we perform both these tests to rigorously test BERT’s knowledge of constructions as positive results for only one could be a result of some other variable.
\bigskip

We make all of our code and experiment details including hyperparameters and pre-trained models publicly available in the interest of reproducibility and so others might experiment with these models.\footnote{\url{https://github.com/H-TayyarMadabushi/CxGBERT-BERT-meets-Construction-Grammar}}

\section{Construction Grammar (CxG)} 
\label{section:cxg}

\begin{figure}[ht]
    \centering
    \includegraphics[width=0.8\textwidth]{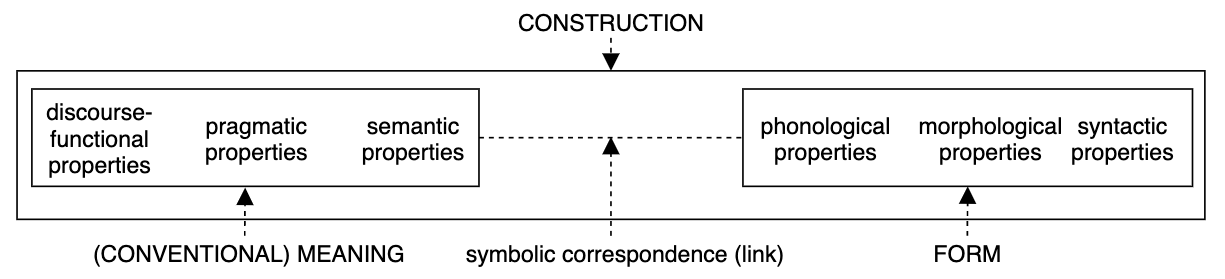}
    \caption{\label{fig:cxg-illustration} Constructions and symbolic pairing of meaning and form. Adapted from ~\protect\cite{croft_cruse_2004}. }
\end{figure}

Construction Grammars are a set of linguistic theories that take constructions as their primary object of study. A construction is a form-meaning pair (see Figure \ref{fig:cxg-illustration} for an illustration and Table \ref{table:cxg-examples} for examples). That is, it has a syntactic and phonological form which has a function or a meaning, albeit a (very) schematic one. Constructions come in various shapes and sizes, from word forms to fully schematic argument structures. The common denominator is that either they are frequently occurring or that their meaning is not predictable from the sum of their parts. A famous example of this is the sentence \textit{She sneezed the foam off the cappuccino}, which is an instance of the \texttt{caused-motion} construction. The verb \textit{sneeze} on its own cannot be interpreted as a motion verb, nor is it usually used as a ditransitive verb, i.e. it does not normally take any complements. It is the caused-motion construction that activates or highlights the motion dimension of the verb ``sneeze''. Other examples of the caused-motion construction include \textit{She pushed the plate out of the way} or \textit{They moved their business to Oklahoma}. All these constructions share a similar syntactic pattern (form) and a meaning of caused motion.

Rather than looking at words as individual tokens ordered in a sentence on the basis of syntactic rules, CxG assumes that syntax has meaning. According to CxG, linguistic knowledge is made up of chunks (i.e. constructions) such as  the partially filled idiom ``drive someone X'', instances of which could be ``drive someone crazy/nuts/bananas'' etc. which is in turn an instance of the resultative construction ``X MAKES Y Z'' e.g \textit{Sarah hammered the metal flat}. CxG assumes that speakers are able to recognise a pattern after having encountered this pattern a certain number of times, with various lexical items. This is similar to merging n-grams (or chunks) through levels of schematicity or abstraction. It is assumed that speakers achieve this level of generalisation by abstracting over a number of similar instances. The examples given in Table \ref{table:cxg-examples} illustrate a lower-level of generalisation where part of the construction is lexically filled, with \textit{didn't} and \textit{how} as fixed elements. The negative auxiliary \textit{didn't} is an instance of the more schematic negative construction ``AUX + \textit{not}'', which is used with most verbs. 

\begin{table}[ht]
\footnotesize
\begin{center}
\renewcommand{\arraystretch}{1.25} 
\begin{tabular}{|L{15.5cm}|}
\hline
\textbf{She didn't understand how} I could do so poorly. \\
Kiedis recalled of the situation: ``He had such an outpouring of creativity while we were making that album that I think \textbf{he really didn't know how} to live life in tandem with that creativity.'' \\
\textbf{We didn't know how} or why. \\
One day she picked up a book and as she opened it, a white child took it away from her, saying \textbf{she didn't know how} to read. \\
In a 1978 interview,  Dylan reflected on the period: ``\textbf{I didn't know how} to record the way other people were recording, and I didn't want to. \\
And it can be on my album, too, \textbf{I just didn't realize how} it worked\dots At first when I got this, people didn't know that I was an artist, so it was, like, `Oh, this songwriter BC.' \\
\hline
\end{tabular}
\end{center}
\caption{\label{table:cxg-examples} Examples of sentences containing the Construction {\small \emph{Personal Pronoun + didn't + V + how}} identified using a modified version of Dunn's~\protect\shortcite{dunn_2017} work (see Section  \ref{section:c2xg}) with the pattern highlighted in bold.}
\end{table}

For example, after encountering several sentences that are instances of the same construction as \textit{She put a finger on that}, children associate the word `put' with the pattern ``X (she) causes Y (finger) to move to Z\_location (that)'', even when the verb `put' itself is not present, as in the sentence \textit{He done boots on.} Similarly, \newcite{KASCHAK2000508} demonstrated that when people encounter words used in novel ways, they rely on constructions to decode their meanings. \textit{She crutched him the ball}, for example, is interpreted to mean that she transferred the ball to him using the crutch, whereas \textit{she crutched him} is interpreted as she hit him with a crutch. This phenomenon explains the human ability to generate and understand the infinite creative potential offered by language~\cite{chomsky1957syntactic} based on finite input.

\section{Related Work}

The study of how BERT works, what it captures and what it is capable of - called BERTology - has been gaining momentum since the introduction of BERT~\cite{DBLP:journals/corr/abs-2002-12327}. Of particular relevance to us is the work associated with \emph{probing}. Probing is the use of a supervised model to  establish if a particular encoding of a sentence contains certain information~\cite{conneau-etal-2018-cram}, such as say PoS information~\cite{belinkov-etal-2017-neural} or sentence length~\cite{DBLP:conf/iclr/AdiKBLG17}. It has been used to establish that BERT seems to capture more detailed information such as the entire classical NLP pipeline~\cite{DBLP:journals/corr/abs-1905-05950} and measure BERT's ability to reason~\cite{DBLP:journals/corr/abs-2005-00782}. 

Probing techniques have also been used to understand the linguistic information captured by contextual representations as in the recent work by ~\newcite{liu-etal-2019-linguistic}. They have also been used to establish that multilingual BERT learns representations of syntactic dependency labels which largely agree with the Universal Dependencies taxonomy~\cite{chi-etal-2020-finding}. Of particular interest to this work is a probing technique called \emph{edge probing}, which consists of tasks, inspired by traditional structured NLP,  designed to evaluate how models encode sentence structure across syntactic, semantic, local and long-range phenomenon ~\cite{tenney2018what}. Some of this work has extended prior work that similarly investigated RNNs, such as whether RNNs can learn to predict subject-verb agreement~\cite{gulordava-etal-2018-colorless,TACL972}.

\subsection{Constructions and Semantic Relatedness}
\label{section:cxg-semantics}

\newcite{jbp:/content/journals/10.1075/ijcl.8.2.03ste} showed that certain verbs are preferentially used in certain constructions, leading to the conclusion that sentences that are instances of the same construction share some semantic relation. The verbs that prototypically occur in one construction are assumed to indicate that construction's archetypal meaning. Examples for the ditransitive constructions would be \textit{give} and \textit{tell}, both of which express the concept of transfer to some extent. Because of this shared meaning, a sentence containing this construction will be interpreted accordingly. Consider the sentences: a) \emph{She sang me a song}, b) \emph{She threw me a bag} and, c) \emph{She blew me a kiss}. These sentences are all instances of the ditransitive construction and share the semantic meaning of X gave Y Z or the idea of ``transfer''. However, it is important to note that out of this context \emph{sang}, \emph{threw}, \emph{blew} are not semantically similar. 

Constructions, like lexical items, are polysemous and can be associated with distinct but related senses, and these constructions themselves have been shown to be interrelated \cite{goldberg1995constructions}. Although Goldberg has argued that there is more to gain by looking at various instances of the same construction, she also acknowledged that speakers are aware of how ``alternating'' constructions are related~\cite{SurfacegeneralizationsAnalternativetoalternations}. Additionally, \newcite{Goldwater2011StructuralPA} showed that when people are ``primed'' with certain constructions, they tend to produce more sentences of those constructions. 

From a computational linguistics perspective,~\newcite{tsao-wible-2013-word} used features from CxGs for semantic similarity but were restricted by the availability of CxG information at the time.

\subsection{Computational Learning of Construction Grammars}
\label{section:c2xg}

The computational generation of construction grammars using distributed semantics is a relatively new area of research and there have been two major studies in this regard. The first is the work by \newcite{dunn_2017} who presents an algorithm - the grammatical induction algorithm - for learning construction grammar from a dataset. The second is work by \newcite{doi:10.1080/09296174.2020.1767481} who similarly extract a pattern grammar (another grammar that aims to account for both the lexicon and syntax or semantics \cite{sinclair1987looking,hunston2000pattern}) from text. Pattern grammar differs from Construction Grammars in that it remains agnostic as to the role played by these patterns in the acquisition and storage of linguistic information in speakers. That is, it does not claim to be cognitively realistic, but is merely descriptive. In this work we adapt Dunn’s \shortcite{dunn_2017} system to identify the constructions that a given sentence instantiates. 

\section{Experimental Set-up}
\label{section:experimental-set-up}

We design two sets of experiments to answer each of our questions: a) How does the addition of constructional information affect BERT and, b) How effective is BERT in identifying constructions? Thus, the first set of experiments (described in Section \ref{section:cxgbert-exp-design}) are aimed at adding constructional information to BERT, testing the resultant model using probing techniques and on downstream tasks and, the second set, described in Section \ref{section:probing} consist of probing standard BERT models using a variety of probes to test how effective BERT is at identifying constructions. Given the nature of our experiments, and the fact that we use non-standard datasets, we establish baselines for each of our tasks and perform an ablation study to better understand the relation between CxGs and BERT.  

For use in both sets of experiments, we classify all sentences in the WikiText-103 corpus, which consists of a subset of verified ``Good'' and ``Featured articles'' on Wikipedia~\cite{DBLP:journals/corr/MerityXBS16} into different ``documents''. This classification is based on the constructions each sentence is an instances of which is extracted using a modified version of Dunn's \shortcite{dunn_2017} work. Their system provides a list of over 22 thousand constructions and we use this pre-calculated list of constructions to classify sentences. It should be noted that a single sentence can be classified as being an instance of several constructions. We modify their system to vastly speed it up so as to process all sentences from WikiText-103, which consists of 30,000 articles and approximately 4.6 million sentences. Examples of some sentences instantiating one construction so processed are listed in Table \ref{table:cxg-examples}.

This results in between 0 and over 50,000 sentences instantiating each of the constructions defined. We call the collection of ``documents'', each consisting of sentences from WikiText-103, that are instances of a particular construction ``CxG WikiText''. Some sentences may not be classified as an instance of any construction so are discarded. Table \ref{table:cxg-examples} provides an illustration of one such ``document''. Since constructions that are instantiated by fewer sentences/are less frequent in our dataset are likely to constrain the meaning enough (not too general) for them to be useful, we further divide constructions based on their frequency (i.e. how many sentences are instances of each construction) in each of our experiments. 

Both sets of experiments are designed based on the premise that there is significant linguistic information encoded in the knowledge that two sentences are instances of the same construction. As discussed in Section \ref{section:cxg-semantics}, there is significant theoretical and applied linguistic evidence to support this claim. This reduces the problem of classifying sentences into over 22 thousand constructions into a binary classification problem of whether or not two sentences are instances of the same construction. 

\subsection{CxGBERT: Encoding Constructions into BERT}
\label{section:cxgbert-exp-design}

The Next Sentence Prediction (NSP) objective that BERT is trained on, which requires the model to predict if one sentence follows another or is from a different document, allows BERT to learn relationships between sentences. To answer the first of our questions: How does the addition of constructional information affect BERT, we make use of the NSP objective by replacing training documents, which in the case of BERT are either Wikipedia articles or book sections from the Book Corpus, with the CxG WikiText (Section \ref{section:experimental-set-up}) thus converting this pre-training objective to a ``same CxG identification'' one. This BERT clone trained on CxG WikiText from scratch for half a million steps will be referred to as``CxGBERT'' (pronounced sig-BERT). 

Since we train CxGBERT on a subset of WikiText-103, we must train another instance of BERT (which we refer to as the BERT Base Clone) on the same data so as to have a comparable baseline. We do this by using the original Wikipedia articles contained in WikiText-103. However, since some sentences are not instances of any construction (as identified by our the model) and therefore not included in CxG WikiText, we remove these from WikiText-103 while also creating an ``article break'' when dropping a sentence. This ensures that consecutive sentences are true next sentences. Finally, sentences are often instances of more than one construction. For example, the sentence \textit{She pushed the books out of the way} is an instance of the \texttt{caused-motion construction `X move Y Z'} (where Z is the path), the somewhat idiomatic construction \textit{out of the way}, the plural constructions \texttt{Noun-s} (e.g., \textit{books}). This implies that these sentences appear in several of the `documents' used for CxGBERT's training data requiring that we make multiple copies of BERT Base Clone’s training data so as to have exactly the same number of training sentences for CxGBERT and BERT Base Clone.

To ensure that any difference in performance between CxGBERT and BERT Base Clone is a result of the ``construction documents'' and not some oddity of the training process, we train a third model on exactly the same data as BERT Base Clone but with the sentences (including document breaks) randomised - we call this BERT Random.

These three BERT models are therefore trained from scratch, using the BERT base architecture, on exactly the same sentences. The only difference is what constitutes a ``document'' in the pre-training data (either sentences clustered based on the CxG they are an instance of, in the case of CxGBERT or a Wikipedia article, in the case of BERT Base Clone), the ordering of the sentences and how often each sentence is repeated. All sentences are repeated an equal number of times in BERT Base Clone and BERT Random, whereas in the case of CxGBERT, a sentence is repeated as often as the number of constructions it is an instance of. 

Additionally, as discussed in Section \ref{section:experimental-set-up}, the number of sentences that are instances of each construction varies drastically between 2 and well over 50,000. To account for this, and to ensure homogeneity amongst constructions, we split the constructions available into those whose frequency is between 2 and 10,000 instances (the Lower set) and those whose frequency is above  10,000 (the Upper set).

This results in six BERT clones: CxGBERT trained using sentences instantiating constructions that have a frequency from 2 to 10,000 instances  (which we call Lower CxGBERT) and its corresponding BERT Base Clone and CxG Random models (which we call Lower BERT Base Clone and Lower BERT Random). We also have a similar set of three BERT clones associated with constructions whose frequency is above 10,000 instances: Upper CxGBERT, Upper BERT Base Clone and Upper BERT Random. 

Finally, we continue pre-training from the BERT Base pre-trained checkpoint, made available by \newcite{DBLP:journals/corr/abs-1810-04805}, with the Lower CxGBERT training data for 20 and a 100 thousand steps. This attempts to make up for the significantly smaller dataset used in pre-training our BERT clones. We hope that by pre-training from a checkpoint that has been training on all of Wikipedia and the Books Corpus we can ``infuse'' CxG information into pre-training models thus making use of the larger training data available to standard pre-trained models and constructional information available to CxGBERT. This results in two additional BERT clones which we call BERT Plus CxG 20K and BERT Plus CxG 100K.

When pre-training the BERT clones described in this section, we use the same hyperparameters as those set out by \newcite{DBLP:journals/corr/abs-1810-04805} with two changes so as to speed up training: we reduce the maximum sequence length to 128 as suggested in the original work but do not perform additional pre-training (for a smaller number of steps) using a sequence length of 512, and also reduce the number of pre-training steps from 1 million to 500 thousand. While these changes (and the different data we use) mean that we cannot compare our BERT clones to the original BERT, the BERT clones remain comparable to each other. Each BERT clone is pre-trained once so we have one version of each clone pre-trained from one random initialisation. Details of these training procedures are included in the associated program code and experimental results we release. 

All \emph{eight} BERT clones are evaluated on a subset of The General Language Understanding Evaluation (GLUE) tasks~\cite{wang-etal-2018-glue},  and SQuAD 1.0~\cite{rajpurkar-etal-2016-squad}. See results in Section \ref{section:encoding-constructions-results}

\subsubsection{Probing CxGBERT}
\label{section:probing-cxgbert}

We  ``probe'' Lower CxG BERT and Lower BERT Base Clone using edge probing~\cite{tenney2018what}. We use a subset of the sub-sentence tasks that are a part of edge probing to discover the differences in the encoding of sentence structure captured by CxG BERT. This is aimed at evaluating how constructional information alters BERT's ability to capture sentence structure. See results in Section \ref{section:probing-cxg-bert}.

\subsection{BERT’s Knowledge of Construction Grammar}
\label{section:probing}

To answer the question: ``How effective is BERT in identifying constructions?'', we create a set of probes that require BERT to predict if two sentences are instances of the same construction. \emph{It is important to note that the version of BERT we use for these experiments is the standard BERT base trained on Wikipedia and the Book Corpus, unadulterated by any constructional pre-training.} We use the BERT Base Cased (cased\_L-12\_H-768\_A-12) for these experiments. Once again, due to the significant variance in the frequency of each construction, we break up the constructions into sets based on their frequency: 2 to 50, 50 to 100, 100 to 10,000, above 10,000 and between 2 and 10,000.

We note that each of the sets of constructions (e.g. those with between 2 and 50 sentences) have a different number of constructions. To ensure that we can compare between these sets, we pick exactly 2 positive and 2 negative pairs from each set for training and 1 positive and 1 negative pair each for the development and test sets. The alternative would have been to pick the same number of training and test samples from each set, but have a different number of examples from each construction. 

An important element of probing is ensuring that the sentence representation provided by an encoder is not fine-tuned (frozen)~\cite{tenney2018what} during the training of the supervised model - this ensures that information specific to the probe does not filter into the sentence representations, forcing the supervised model to learn the mapping between the representation of information pertaining to this probe (e.g. PoS information) within the embedding and the probing task (e.g. PoS tagging). 

However, in freezing BERT, we might not be capturing information contained within BERT's internal layers as it might not be explicitly expressed in the output vector we choose to use (such as the vector representing [CLS] corresponding to the second last layer). Additionally, the internal attention weights of BERT might also carry important information. To get around this, we adapt a variation of the probing strategy proposed by \cite{richardson2020probing} who themselves adapt a version of \emph{inoculation by fine-tuning}~\cite{liu-etal-2019-inoculation}. Inoculation by fine-tuning was originally aimed at testing if challenge datasets (i.e. datasets aimed at testing how brittle models trained on existing benchmarks are), were ``difficult'' for a model because they truly capture phenomena that models cannot capture or because of limitations of the training set. It consists of exposing the model to a ``small'' amount of training data from the new dataset. Unlike \newcite{richardson2020probing}, however, we do not fine-tune BERT on any task other than the CxG task. 

Given these restrictions we test BERT's ability to distinguish between sentences belonging to the same construction and those that do not (in a sense testing BERT's knowledge of constructions) using the following strategies: a) Without fine-tuning BERT whatsoever, b) by freezing the transformer layers of BERT and using 7 fully connected layers on top (we establish this and other relevant hyperparameters using an independent development set as described in the associated code and data), c) inoculating BERT with 100, 500, 1000, and 5000 training examples and d) training it on the full training data (consisting of 2 positive and 2 negative examples from each construction). 

Since \newcite{dunn_2017} uses distributional semantics to find constructions, we must ensure that the resultant constructions do not retain this information to an extent that can be captured by a simpler model. To this end we add a final experiment that consists of measuring the effectiveness of a GloVe based biLSTM model on the same CxG disambiguation task when trained on the entire training set. 

Results of the experiments described in this section, using each of the above probes to test BERT's ability to distinguish constructions (and so its knowledge of constructions) are presented in Section \ref{section:berts-knowledge-of-cxg}. 

\section{Empirical Evaluation}

This section details the results of the experiments described in Experimental Set-Up (Section \ref{section:experimental-set-up}). We do not lowercase any of the input, either during evaluation, probing or fine-tuning and also use the upper case version of BERT base when using pre-trained models. Additionally, the results are the maximum of five runs to account for variation due to random initialisation except where otherwise stated.

\subsection{CxG BERT: Encoding Constructions into BERT}
\label{section:encoding-constructions-results}

We present the evaluation of the various BERT clones on the development sets of a subset of the Glue tasks and SQuAD 1.0 along with their accuracy and loss on the Masked Language (ML) and Next Sentence Prediction objectives (NSP) in Table \ref{table:cxg-on-glue}. 

\begin{table}[ht]
\footnotesize
\def\arraystretch{1.1}
\centering
\begin{tabular}{|p{1.43cm}|p{1.15cm}p{1.15cm}p{1.15cm}|p{1.15cm}p{1.15cm}p{1.15cm}|p{1.15cm}p{1.15cm}p{1.15cm}|}
\hline
  & \textbf{Lower BERT Base Clone} & \textbf{Lower CxG BERT} & \textbf{Lower BERT Random} & \textbf{Upper BERT Base Clone} & \textbf{Upper CxG BERT} & \textbf{Upper BERT Random} & \textbf{BERT Base} & \textbf{BERT Plus CxG 20K} & \textbf{BERT Plus CxG 100K}  \\
 \hline 
 &&&&&&&&&\\[-2ex]
 \hspace{-0.3em}\textbf{ML Acc} & 0.7751& 0.7632& 0.6957& 0.7370& 0.6953& 0.6805& N/A & 0.5971 & 0.6202 \\
 \hspace{-0.3em}\textbf{ML Loss} & 1.0065& 1.0638& 1.3717& 1.1854& 1.4242& 1.4833& N/A & 2.1308 & 1.9310 \\
 \hspace{-0.3em}\textbf{NSP Acc} & 1.0000& 1.0000& 0.9913& 1.0000& 0.9938& 0.9700& N/A & 0.7963 & 0.8488 \\
 \hspace{-0.3em}\textbf{NSP Loss} & 1.24E-5& 4.88E-5& 1.78E-2& 8.08E-6& 2.33E-2& 9.33E-2& N/A & 4.20E-1 & 3.11E-1 \\[1ex]
 \hline
 &&&&&&&&&\\[-2ex]
 \hspace{-0.3em}\textbf{MRPC} & 83.1210& \textbf{84.0391}& 80.9892& \textbf{83.9428}& 81.8040& 81.4241& \textbf{87.7551} & 85.8553& 86.0884 \\
 \hspace{-0.3em}\textbf{CoLA} & 36.3602& \textbf{38.3863} & 30.8298& \textbf{40.9991}& 34.3342& 37.6885& 57.7848&  56.7550& \textbf{59.5925} \\
 \hspace{-0.3em}\textbf{STS} & \textbf{80.9079}& 79.8130& 28.1752& \textbf{83.5393}& 65.2152& 29.5565& \textbf{88.5490}& 87.4546& 86.9574 \\
 \hspace{-0.3em}\textbf{RTE} & \textbf{60.2888} & 56.6787& 50.1805& \textbf{64.2599}& 54.8736& 55.5957& 62.8159& \textbf{64.6209}& 58.8448 \\
 \hspace{-0.3em}\textbf{SST} & \textbf{91.2844}& 89.9083& 89.9083& \textbf{91.2844}& 91.1697& 90.5963& 91.9725& 92.3165& \textbf{92.4312} \\[1ex] 
 \hspace{-0.3em}\textbf{SQuAD 1.0} & \textbf{77.8262}& 76.2176& 73.2891& \textbf{81.9994}& 75.9331& 71.9489& \textbf{85.3970}& 84.8543& 84.7005 \\[1ex]
 \hline
\end{tabular}
\caption{\label{table:cxg-on-glue}Evaluation of BERT clones on the development sets of some Glue tasks and SQuAD 1.0 and their performance on the ML and NSP objectives. We report F1 scores for MRPC, Spearman correlations for STS-B, Matthews Correlation Coefficient for CoLA, and accuracy scores for the other tasks. Maximum scores for each set are highlighted in bold along each row. All models are fine-tuned using hyperparameters listed by \protect\newcite{DBLP:journals/corr/abs-1810-04805}. }
\end{table}

Our results clearly show that how sentences are clustered into documents has a significant impact, as can be seen by the low performance of both random baselines. The fact that the Upper BERT Base Clone performs better than the Lower BERT Base Clone can be attributed to the fact that there are more sentences in the Upper training data, and indeed why we require two independent baselines. Upper CxGBERT consistently fails to beat the Upper BERT Base Clone suggesting that there is possibly less information contained in the constructions that are more frequent. 

The largest gap between BERT Random and the other two models is in the Semantic Text Similarity (STS) task which is aimed at testing if two sentences mean the same, suggesting that the NSP objective helps in capturing semantic similarity regardless of whether sentences are clustered based on Wikipedia articles or constructions. We note that corpora on which we improve upon BERT Base when it is further pre-trained with CxG information, namely The Corpus of Linguistic Acceptability (CoLA) and the Stanford Sentiment Treebank (SST), are both \emph{single sentence} classification corpora, which is surprising as we alter BERT's NSP objective. 

The relative close performance of CXGBERT and BERT Base Clone suggests that constructional information and topical information (as extracted from documents pertaining to a single Wikipedia article) seem to be very similar as measured by performance on downstream tasks. 

\subsubsection{Edge Probing CxGBERT}
\label{section:probing-cxg-bert}

\begin{table}[!htbp]
\setlength\dashlinedash{0.2pt}
\setlength\dashlinegap{1.5pt}
\setlength\arrayrulewidth{0.3pt}
\footnotesize
\def\arraystretch{1.1}
\centering
\begin{tabular}{|L{2.3cm}| C{2.3cm}|C{2.3cm}|C{2.3cm}|C{2.3cm}|}
\hline
\textbf{Probe} & \textbf{Lower BERT Base Clone F1} & \textbf{Lower CxGBERT F1} & \textbf{$\Delta$ CxG Base} & \textbf{$\Delta$ BERT ELMo} \\
\hline
\textbf{PoS} & 0.807& 0.807& 0.0& 0.0 \\
\textbf{NER} & 0.829& 0.844& 1.5& 0.6 \\
\textbf{SRL} & 0.687& 0.666& -2.1& 1.2 \\
\textbf{SPR} & 0.786& 0.784& -0.2& 0.7 \\
\textbf{Non-terminal} & 0.609& 0.600 & -0.9& 2.1 \\
\hdashline
\textbf{NER Task} & 0.7022 & 0.7004 & -0.18 & N/A \\
 \hline
\end{tabular}
\caption{\label{table:probe}Results of Edge Probing on the Lower BERT Base Clone and Lower CxGBERT. The probing tasks are PoS, Named Entity Recognition (NER), Semantic role labelling (SRL), Semantic proto-role (SPR), and non-terminal constituents. $\Delta$ BERT ELMo represents BERT's improvement over ELMo as reported by \protect\newcite{tenney2018what} provided here for reference. Since CxGBERT does better on the NER probe, we include results on the CoNLL-2003 NER task in the row labelled ``NER Task''.Edge probes were run three (not five) times each with a different initialisation.}
\end{table}

We present the results of edge probing Lower BERT Base Clone and Lower CxGBERT in Table \ref{table:probe}. We find that the two models are comparable, with CxGBERT performing better on one, BERT Base Clone on one and the two performing close (albeit with BERT Base Clone performing better) on the other three probes. To test if we can translate CxGBERT's improved performance on the Named Entity Recognition probe into better performance on a downstream task, we test the two models on the CoNLL-2003 NER task. Surprisingly CxGBERT performs worse on the task despite performing better on the probe. We conclude that the two models' ability to capture sentence structure is comparable. 

\subsection{BERT’s knowledge of Construction Grammar}
\label{section:berts-knowledge-of-cxg}

The accuracy of BERT Base (with no additional pre-training) on the CxG disambiguation task is presented in Table \ref{table:bertie}. These results seem to suggest that BERT has a surprising amount of constructional information, being able to predict if two sentences are instances of the same construction with an accuracy of close to 90\% after training on just 500 examples. More training data seems to result in less local minima (as measured by the number of runs that reach ``close'' to the maximum accuracy). The relatively low performance of the GloVe baseline seems to suggest that the constructions used are not too semantically similar despite the use of distributional semantics in generating them. Finally, the gap in performance between the frozen version of BERT and the inoculated versions seems to suggest that BERT captures constructional information in its internal structure. 

\begin{table}[!ht]
\footnotesize
\def\arraystretch{1.1}
\centering
\begin{tabular}{|C{1.6cm}| R{0.8cm}|p{1.2cm}|p{1.0cm}|p{1.0cm}|p{1.0cm}|p{1.0cm}|p{1.0cm}| p{1.2cm}| p{1.2cm}|}
\hline
 & \textbf{\# of CxGs} & \textbf{Full Training} & \textbf{Inoc 5000} & \textbf{Inoc 1000}& \textbf{Inoc 500} & \textbf{Inoc 100} & \textbf{Freeze} & \textbf{No Training} & \textbf{GloVe biLSTM Baseline} \\
 \hline
 \textbf{2-50} & 6384& 95.0501& 92.8571& 90.8130& 88.9333& 72.5564& 73.5119& 63.9646& 69.3163 \\
 \textbf{50-100} & 2696& 93.6573& 92.7114& 90.4488& 88.9466& 70.7901& 70.8828& 63.9651& 62.1732 \\
 \textbf{100-1000} & 8974& 94.4451 & 91.8041 & 89.4974 & 86.5612 & 60.9594 & 68.2249 & 58.8478 & 53.9919 \\
 \textbf{1000-10000} & 3266& 89.7734& 87.8292& 83.2364& 69.7336& 57.7771& 60.9461& 57.7924& 54.1405 \\ [1ex]
 \textbf{$<$10000} & 21216 & 94.4075& 90.1772& 88.6619& 85.7325& 68.3352& 69.3038& 61.1708& 60.8819 \\  
 \textbf{$>$10000} & 465 & 72.5498& 72.5100& 64.9004& 54.9402& 53.3865& 54.7809& 52.8685& 52.1306 \\ [1ex]
 \textbf{All} & 21681 & 93.4851& 89.3409& 87.1339& 85.8009& 64.1137& 69.0905& 60.1541& 62.4242 \\ 
 \hline
\end{tabular}
\caption{\label{table:bertie} Evaluation of BERT Base (with no additional pre-training) on predicting if two sentences are instances of the same construction. Freeze represents the results of BERT frozen, Inoc the number of training samples used for inoculation (Section \ref{section:probing}), and \# of CxGs the number of constructions contained in that set. Rows represents the count of sentences that are instances of constructions in that set.}
\end{table}

\section{Discussion}

In an attempt to determine to what extent BERT has access to constructional information, we set out to answer two questions, the first of which was: How does the addition of constructional information affect BERT? Overall, our experiments show that the addition of constructional information has little impact on BERT. The fact that neither CxGBERT nor BERT Base Clone comes out ahead shows that the information available to both models is similar. However, it is interesting that the linguistic information available to a generalisation engine (such as a neural network or person), regardless of whether the input is a set of Wikipedia articles (as in the case of BERT Base Clone) or sentences instantiating the same construction (as in the case of CxGBERT), is very similar. One need only study the diverse set of sentences that can be instances of the same construction (Table \ref{table:cxg-examples}) to see that this is a non-trivial outcome. This outcome possibly results from the fact that similar concepts are expressed in similar ways (as distributional analysis has shown) so when talking about a given topic, one is likely to be using a limited set of words which trigger a limited set of constructions. Additionally, the fact that BERT Random consistently performs poorly shows that the performance gain is indeed coming from the clustering of sentences in the training data. 

The second question we aimed to answer was: How effective is BERT in identifying constructions? We find that the constructions that are most easily predicted by BERT (using the same CxG prediction task) also tend to have a lower number of sentence instances. A manual inspection of these constructions revealed that these constructions (that have fewer sentences as instances), such as \texttt{Personal Pronoun + didn't + V + how} illustrated in Table \ref{table:cxg-examples} also constrain the meaning of the construction more. Those that have several sentences as instances, such as \texttt{Noun -s}, which BERT finds harder to predict, tend to be so general that they are less useful. The extremely high accuracy on the set of constructions that have less than 10,000 sentences as instances seems to suggest that BERT has a substantial amount of information pertaining to semantically specific constructions and the fact that BERT, when frozen, performs rather poorly seems to suggest that this information is contained within the internal layers of BERT and not in its output. It is striking that these constructions can be predicted by BERT with an 85\% accuracy after training on just 500 training elements. This is particularly surprising given that there are over 21 thousand constructions which are contained in this set - so 500 training elements are not even enough to provide one sample from each construction. 

The combination of these results allows us to conclude that BERT does indeed have access to a significant amount of constructional information. This information, as with other information such as PoS information or parse trees, is not explicitly available in the output layer, but can be accessed from within the internal layers of BERT. The impact of this observation is potentially far reaching as it not only further shows the capabilities of deep learning methods, but also shows that information that is typically called constructional can be learned form exposure to lexico-semantic information. This is expected given the redundancy inherent in language: words and constructions constrain each other mutually. 

We note that this work is limited by the constructions that are available to us. A manual inspection of the constructions seems to indicate that some constructions - even those that have few sentences as instances - seem to be relatively short, even if the sentences that contain them are themselves long. For example, one of the constructions extracted is \texttt{MODAL + BE + Past-Participle} and sentences that are instances of this construction include a) \textit{These complexes, though their origins \underline{may be found} as early as the 19\textsuperscript{th} century, snowballed considerably during the Cold War.} and b) \textit{When the counter-revolution became stronger Moreno called the Junta and, with support from Castelli and Paso, proposed that the enemy leaders \underline{should be shot} as soon as they were captured instead of brought to trial.} However, we also note that verifying if a set of sentences are instances of a higher level construction is cognitively demanding, and so it is possible that we were only able to identify the shorter, more obvious, patterns. This might be contributing to BERT’s high accuracy in predicting them, yet does not fully explain BERT’s ability to predict similarity amongst over 21 thousand constructions using just 500 training examples. 

\section{Conclusions and Future Work}

In this work, we set out to explore the extent to which BERT might have access to constructional information by use of several probes and extensive experimentation. Our results allow us to conclude that BERT does indeed have access to a significant amount of information, much of which linguists typically call constructional information. We hope that this work will inspire greater interaction between neural network research and linguistics,  as has been suggested before~\cite{DBLP:journals/corr/abs-1809-04179}.

An analysis of constructions that have several sentences as instances showed that they tend to contain generic labels (such as ``Preposition + his'') and do not constrain the meaning of the construction or its slots much. It is doubtful whether such patterns would be considered constructions on a Construction Grammar approach, since these patterns have form, but lack a clear mapping of that form to a specific function or meaning. Dunn \shortcite{dunn_2017} describes a method of filtering constructions that might address this issue and we believe it to be an interesting future direction of work.

Finally, these results also suggest that BERT could be used by linguists for creating an inventory of constructions. There have so far been some attempts at creating such an inventory, usually called a constructicon~\cite{10.5555/1867135.1867182}, but the tools used for these endeavours do not match CxGBERT's potential~\cite{lyngfelt2018constructicography}. For example, rather than a purely manual analysis, or a mere frequency split, one could use BERT's ability to disambiguate a construction as indicative of how ``informative'' it is, yet another possible direction of future work.

\section*{Acknowledgements}

Research supported with Cloud TPUs from Google's TensorFlow Research Cloud (TFRC).

\bibliographystyle{coling}
\bibliography{coling2020}

\end{document}